\documentclass[letterpaper, 10 pt, conference]{ieeeconf}  %

\IEEEoverridecommandlockouts                              %

\usepackage{graphicx}
\usepackage{amsmath}
\usepackage{amssymb}
\usepackage{booktabs}
\usepackage{lipsum} 
\usepackage{multirow}
\usepackage{threeparttable}
\usepackage{tabularx}
\usepackage{tikz}
\usepackage{comment}
\usepackage{dsfont}
\usepackage{color, colortbl}
\usepackage{threeparttable}
\usepackage{tabularx}
\usepackage{caption}
\usepackage{subcaption}
\usepackage{wrapfig}
\usepackage{hhline}
\usepackage{sidecap}

\usepackage{algorithm}
\usepackage{listings}
\usepackage{xcolor}

\usepackage{footnote}
\usepackage{soul}
\usepackage{url}
\makesavenoteenv{algorithm}

\usepackage{hyperref}

\definecolor{VisGreen}{RGB}{1, 128, 0}
\definecolor{linkcolor}{RGB}{237,3,140} 
\newcolumntype{C}{>{\centering\arraybackslash}X}

\hypersetup{
    colorlinks=true, %
    urlcolor=linkcolor    %
}

\title{\LARGE \bf
WidthFormer: Toward Efficient Transformer-based \\ BEV View Transformation
}

\author{Chenhongyi Yang$^{1}$ Tianwei Lin$^{2}$ Lichao Huang$^{2}$ Elliot J. Crowley$^{1}$\\ 
$^{1}$School of Engineering, University of Edinburgh~~~ $^{2}$Horizon Robotics
}

\begin{document}

\maketitle
\thispagestyle{empty}
\pagestyle{empty}

\newcommand{\ourmethod}{WidthFormer}
\newcommand{\ourmethodFull}{Width Former Full}

\newcommand{\ourpe}{RefPE}
\newcommand{\ourpeFull}{Reference Positional Encoding}

\newcommand{\refineFormer}{Refine Transformer}

\begin{abstract}

We present \textit{\ourmethod}, a novel transformer-based module to compute Bird's-Eye-View (BEV) representations from multi-view cameras for real-time autonomous-driving applications.
\ourmethod~ is computationally efficient, robust and does not require any special engineering effort to deploy. We first introduce a novel 3D positional encoding mechanism capable of accurately encapsulating 3D geometric information, which enables our model to compute high-quality BEV representations with only a single transformer decoder layer. This mechanism is also beneficial for existing sparse 3D object detectors. Inspired by the recently proposed works, we further improve our model's efficiency by vertically compressing the image features when serving as attention keys and values, and then we develop two modules to compensate for potential information loss due to feature compression. Experimental evaluation on the widely-used nuScenes 3D object detection benchmark demonstrates that our method outperforms previous approaches across different 3D detection architectures. More importantly, our model is highly efficient. For example, when using $256\times 704$ input images, it achieves 1.5 ms and 2.8 ms latency on NVIDIA 3090 GPU and Horizon Journey-5 computation solutions. Furthermore, \ourmethod~also exhibits strong robustness to different degrees of camera perturbations. Our study offers valuable insights into the deployment of BEV transformation methods in real-world, complex road environments. Code is available at \url{https://github.com/ChenhongyiYang/WidthFormer}.

\end{abstract}

\section{Introduction}

In recent years, the field of vision-based Bird's-Eye-View (BEV) 3D object detection has garnered significant interest and witnessed substantial advancements~\cite{philion2020lift,huang2021bevdet,bevformer,li2022bevdepth,chen2022graph}. In contrast to directly detecting objects from image features, identifying 3D objects from a unified BEV representation aligns more intuitively with human perception and can be readily adapted to tasks such as 3D object tracking~\cite{yin2021center} for autonomous driving. Furthermore, BEV representations can be easily integrated with other modalities like LIDAR point clouds~\cite{li2022bevdepth,chen2023bevdistill}.

\begin{figure*}[!t]
    \centering
    \includegraphics[width=\textwidth]{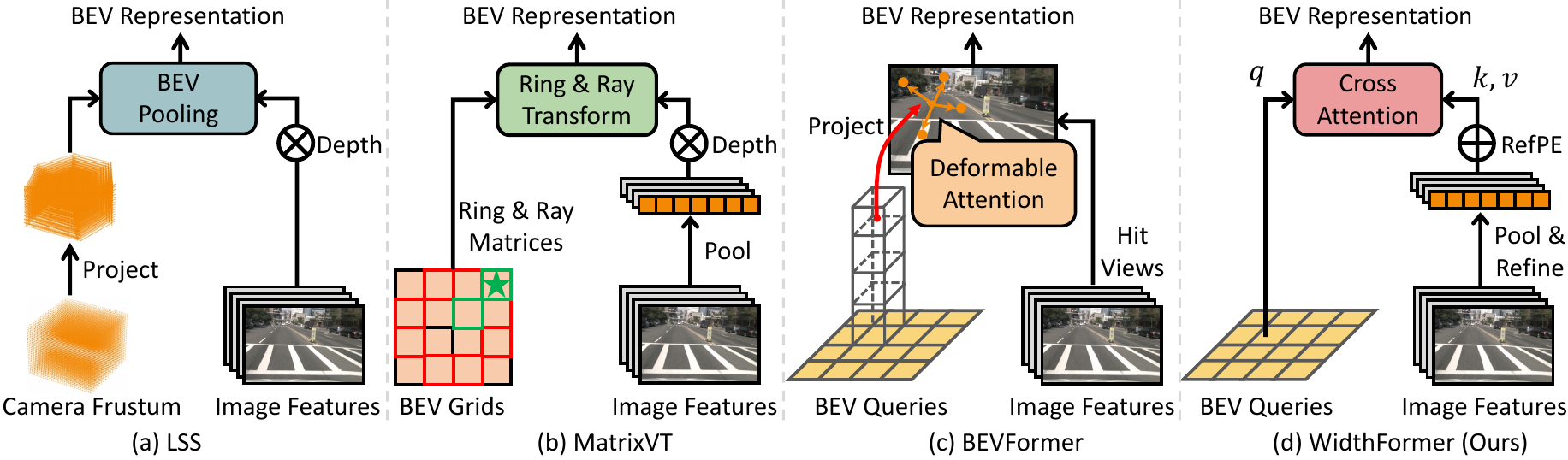}
    \captionsetup{font={small}}
\caption{Comparison on different BEV view transformation paradigms.}
\label{fig:vtCompare}
\vspace{-5mm}
\end{figure*}

The transformation of multi-view image features into a unified BEV representation, also referred as View Transformation (VT), plays a centre role in BEV-based 3D object detection. These can be broadly categorised into two streams: 1) Lift-Splat based approaches~\cite{philion2020lift,huang2021bevdet,huang2022bevdet4d,li2022bevdepth}, which first lift the image features to 3D space, and then gather the projected image features in the vertical direction to yield features for each BEV bin; and 2) Transformer-based methods~\cite{bevformer,jiang2022polar}, which derive BEV representations by querying the image features via attention operations. While both types of methods have achieved great success~\cite{caesar2020nuscenes,sun2020scalability}, their deployment to real-time autonomous-driving applications running on edge-computing devices~\cite{GUTIERREZZABALLA2023102878,j5_website} presents considerable challenges. Here, we outline the three main obstacles that impede the successful deployment of these methods. (1) \textit{Need for non-standard Operations}: Both Lift-Splat-based and Transformer-based methods necessitate special operations that require significant engineering effort for efficient implementation~\cite{huang2021bevdet,liu2022bevfusion,j5_website}. For example, in BEVFusion~\cite{liu2022bevfusion}, a complex CUDA multi-threading mechanism was devised to gather point features in each spatial grid. The deformable attention operation~\cite{deformableDETR} employed by BEVFormer~\cite{bevformer} and PolarFormer~\cite{jiang2022polar} also necessitates specialised engineering effort when being deployed to edge computing devices. (2) \textit{Heavy Computation}: Previous Transformer-based methods~\cite{bevformer,jiang2022polar} utilise multiple layers of transformer decoders to compute the BEV representation, which significantly impacts processing speed and impedes their deployability for real-time applications. Moreover, the stacking of deformable attention operations results in a large amount of random memory reads, which is often a bottleneck for edge-computing devices. (3) \textit{Lack of Robustness}: The camera poses on a vehicle are usually perturbed due to a variety of factors, such as collisions or wear and tear. As we will demonstrate in Sec.~\ref{sec:exp}, such perturbations substantially degrade the quality of BEV representations produced by prior Lift-Splat-based and Transformer-based methods. This makes these methods difficult to apply in real-world, complex driving environments. Given these considerations, we aim to develop a new BEV transformation method that is efficient, robust, and does not necessitate specialized engineering effort for deployment.

In this paper, we introduce~\emph{\ourmethod}: a new transformer-based BEV view transformation method. Our efficient VT module comprises a single layer of the transformer decoder~\cite{Transformer_NIPS2017_Vaswani} and does not involve any non-standard operations. Empowered by~\emph{\ourpeFull~(\ourpe)}, a new method to compute positional encoding for transformer-based 3D object detection. \ourmethod~computes BEV representation by directly querying the image features using the BEV queries computed from the coordinate on the BEV plane. To further mitigate the significant computational cost caused by the large amount BEV query vectors, we follow recently proposed approaches~\cite{zhou2022matrixvt} to compress the image features in the vertical direction, significantly enhancing \ourmethod's efficiency and scalability. In Fig.~\ref{fig:vtCompare}, we show a high-level comparison between \ourmethod~and other BEV transformation methods. Moreover, to compensate for any potential information loss in these compressed features, we designed two techniques: (1) We design an efficient Refine Transformer, where the compressed features attend to and extract valuable information from the original image features, and (2) We propose to train the model with complementary tasks that directly inject task-related knowledge into the compressed features.

To summarise, we make the following four contributions: \textbf{(1)} We introduce \ourmethod, a light-weighted and deployment-friendly BEV transformation method that employs a single layer of a transformer decoder to compute BEV representations. \textbf{(2)} We propose \ourpeFull~(\ourpe), a new positional encoding mechanism for 3D object detection, to help \ourmethod~assist \ourmethod's view transformation. It can also be used to boost the performance of sparse 3D object detectors in a plug-and-play way.  \textbf{(3)} We evaluate the proposed modules on the widely-used nuScenes 3D object detection dataset~\cite{caesar2020nuscenes}. The results show that \ourpe~can significantly improve the performance of sparse object detectors. Also, \ourmethod~outperforms previous BEV transformation methods in both performance and efficiency across a range of 3D detection architectures.

\section{Related Work}

\begin{figure*}[!t]
    \centering
    \includegraphics[width=\textwidth]{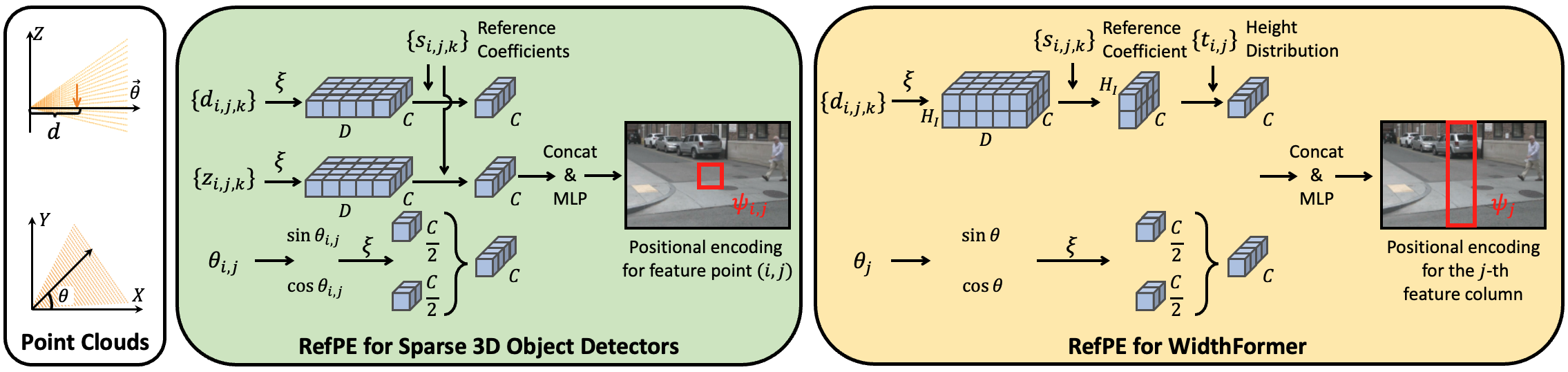}
    \vspace{-0.3cm}
    \captionsetup{font={small}}
\caption{\small \ourpeFull~(\ourpe): \ourpe~has a rotation and a distance part. For sparse 3D detectors, it has another height part. The rotation encoding is computed using a camera ray's rotation degree on the BEV plane. To compute the point-wise distance PE and height PE, we leverage the reference coefficients, predicted from the visual features, to aggregate the distance \& height PEs of reference points on a camera ray. We remove the height PE for width features  and compute their distance PE by aggregating all point-wise distance PE along an image column using a predicted height distribution. }
\vspace{-5mm}
\label{fig:pe}
\end{figure*}

\noindent\textbf{Vision-based 3D Object Detection.}
Image-based 3D object detection is the fundamental module for many downstream applications. Monocular 3D detection~\cite{chen2016monocular,ding2020learning,wang2022probabilistic} aims to detect 3D objects from a single input image. For example,  FCOS3D~\cite{wang2021fcos3d} extends the 2D FCOS~\cite{tian2019fcos} detector for 3D detection by regressing 3D bounding boxes. Multi-view 3D object detection~\cite{chen2022graph,sparse4d,wang2022detr3d} incorporates multiple images for better geometric inference. For example, PETR~\cite{liu2022petr} extend the sparse detector DETR~\cite{detr} by introducing 3D positional encoding. PETRv2 improves PETR~\cite{liu2022petrv2} by incorporating temporal modeling. StreamPETR~\cite{wang2023exploring} proposed a novel query propagation algorithm to better leverage temporal information in a long range. CAPE~\cite{Xiong2023CAPE} improves the positional encoding of PETR by creating positional encodings under local camera coordinate frames. The recently proposed 3DDPE~\cite{shu20223d} incorporates multi-modal supervision to accurately depth, based on which the point-wise positional encodings are computed. BEV-based 3D object detectors~\cite{li2022bevdepth,huang2021bevdet,huang2022bevdet4d,li2023fastbev,zhou2022matrixvt} first transform multi-view images to a unified BEV representation from which 3D objects are detected. BEVDet~\cite{huang2021bevdet,li2022bevdepth} and its follow-up work use LSS~\cite{philion2020lift} to compute BEV features and predict objects using convolutional heads~\cite{yin2021center}. BEVFormer~\cite{bevformer} computes BEV features using deformable attention operations~\cite{deformableDETR} and relies on a DETR-style head~\cite{detr} for object detection.

\noindent\textbf{Vision-based BEV Transformations.}
The intuitive IPM-based methods~\cite{8814050,li2023fastbev} compute BEV features through 3D-to-2D projection and interpolation. A problem with these is that the quality of BEV features will be severely harmed when the flat-ground assumption does not hold. In Lift-Splat~based methods~\cite{philion2020lift,huang2021bevdet,huang2022bevdet4d}, BEV features are computed by vertically pooling the projected point cloud features, weighting by their predicted depth. While being high-performing, the Lift-splat process~\cite{philion2020lift} is far from efficient. BEVFusion~\cite{liu2022bevfusion} accelerates this pooling process with a multi-threading mechanism. M$^2$BEV~\cite{xie2022m} saves memory usage by assuming a uniform depth distribution. MatrixVT~\cite{zhou2022matrixvt} improves overall efficiency by compressing the visual features in the vertical dimension and then computes BEV features using an efficient polar coordinate transformation. BEVDepth~\cite{li2022bevdepth} incorporates point clouds for improved depth estimation. Transformer-based VT methods directly output BEV representations through the attention mechanism.  
CVT~\cite{zhou2022cross} and PETR~\cite{liu2022petr} rely on 3D positional encodings to provide the model with 3D geometric information. To improve efficiency, many recent approaches~\cite{bevformer,jiang2022polar} adopt deformable attention~\cite{deformableDETR}.%

\section{Method}

In this section, we introduce our method in detail.  We introduce the new 3D position encoding mechanism in Sec.~\ref{sec:pe}. Then we introduce our BEV transformation module in Sec.~\ref{sec:bevTransformer}. In Sec.~\ref{sec:refinement}, we describe how we refine the compressed features against potential information loss.

\subsection{\ourpeFull~(\ourpe)}
\label{sec:pe}

In this work, we design a new 3D positional encoding mechanism for both transformer-based 3D object detectors, e.g., PETR~\cite{liu2022petr}, and our BEV view transformation module. Intuitively, for each visual feature that serves as the keys and values in the transformer, we aim to learn its positional encoding by referring to a series of reference points; therefore, we call our method \ourpeFull~(\ourpe). Given the multi-view image features $\mathbf{F}^I \in \mathbb{R}^{N_c \times H_I \times W_I \times C}$, where $N_c$ is the number of cameras; $H_I$, $W_I$ and $C$ are the height and width, and channel dimension, the computed positional encodings $\mathbf{\boldsymbol{\psi}} \in \mathbb{R}^{N_c \times H_I \times W_I \times C}$ has a same shape with the input features. Specifically, given a feature pixel\footnote{Here we omit the view index for simplicity.} whose coordinate on the image plane is $\mathbf{P}_{i,j}=[u_{i,j},v_{i,j}]^{\top}$, we first lift the pixel by associating it to $D$ discrete depth bins, resulting in $D$ reference points whose homogeneous coordinates are $\{ \mathbf{\hat{P}}_{i,j,k} = [u_{i,j}\times d_k, v_{i,j} \times d_k, d_k]^{\top} |~k \in |D| \}$. We project all $D$ points to a unified 3D space by:
\begin{align}
    \mathbf{C}_{i,j,k} = \mathbf{R}^n \cdot \mathbf{I}^{-1} \cdot \mathbf{\hat{P}}_{i,j,k} + \mathbf{T}^n\label{eq:extrinsic},
\end{align}
where $\mathbf{C}_{i,j,k} = [x_{i,j,k}, y_{i,j,k}, z_{i,j,k}]^{\top}$ is the point's 3D Cartesian coordinate; $\mathbf{I} \in \mathbb{R}^{3\times3}$ is the camera intrinsic matrix; $\mathbf{R}^{n} \in \mathbb{R}^{3\times3}$ and $\mathbf{T}^{n} \in \mathbb{R}^{3\times1}$ are the rotation matrix and translation vector that transform the coordinate in the $n$-th view to the unified LIDAR coordinate frame~\cite{caesar2020nuscenes,mmdet3d2020}. 

\begin{figure*}[!t]
    \centering
    \includegraphics[width=0.9\textwidth]{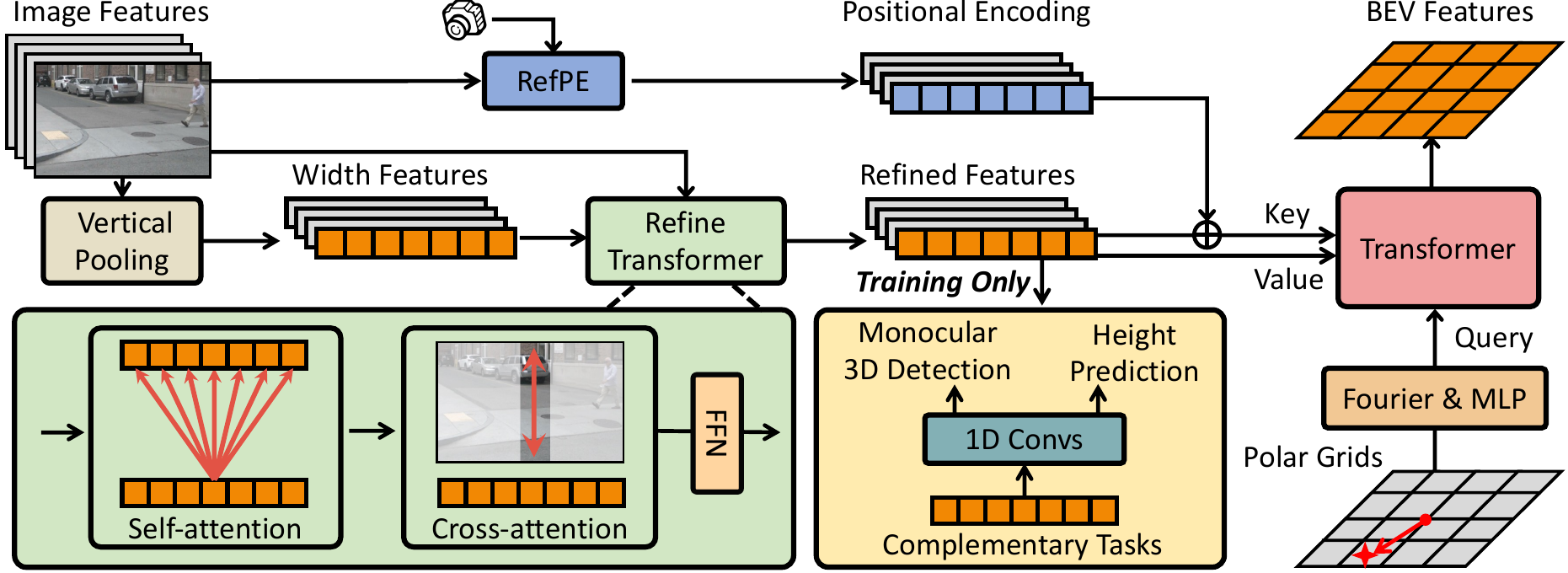}
    \captionsetup{font={small}}
\caption{\small \ourmethod~takes multi-view images as input and outputs the converted BEV features. It first compresses the image features into width features by pooling their height dimension. Then the width features are refined with a Refine Transformer to compensate for potential information loss. After adding with our \ourpeFull, the width features are fed into the transformer decoder to serve as keys \& values, which are queried by the BEV query vectors computed from the pre-defined BEV polar coordinates.}
\vspace{-4mm}
\label{fig:pipeline}
\end{figure*}

We then compute positional encodings using polar coordinates where the ego vehicle is located at the origin: each 3D point's positional information is encoded as the combination of three parts:  1) distance to the ego on the BEV plane, 2) rotation related to the ego on the BEV plane, and 3) height to the ground. Specifically, for the 2D point $\mathbf{P}_{i,j}$ and its reference points $\{\mathbf{C}_{i,j,k}|~k \in |D| \}$, we first compute the positional encodings for all its reference points $\{\tilde{\boldsymbol{\psi}}_{i,j,k}\}$:
\resizebox{1.0\linewidth}{!}{
  \begin{minipage}{1.2\linewidth}
\begin{align}
    &\tilde{\boldsymbol{\psi}}_{i,j,k} = \mathrm{Concat}\Bigl( \xi(d_{i,j,k}), \xi(\sin{\theta_{i,j,k}}), \xi(\cos{\theta_{i,j,k}}), \xi(z_{i,j,k}) \Bigr) \\
    &d_{i,j,k} = \sqrt{x_{i,j,k}^2 +y_{i,j,k}^2},~\sin{\theta_{i,j,k}} = \frac{y_{i,j,k}}{d_{i,j,k}},~\cos{\theta_{i,j,k}} = \frac{x_{i,j,k}}{d_{i,j,k}}
\end{align}
\vspace{1mm}
\end{minipage}}
where $\xi$ is the Fourier positional encoding~\cite{Transformer_NIPS2017_Vaswani}. The positional encoding  of $\mathbf{P}_{i,j}$ is computed by aggregating $\{\tilde{\boldsymbol{\psi}}_{i,j,k}\}$ with a series of reference coefficients and an MLP:
\resizebox{1.0\linewidth}{!}{
  \begin{minipage}{1.1\linewidth}
\begin{align}
    \boldsymbol{\psi}_{i,j} = \mathrm{MLP}\Bigl( \sum_{k=1}^{D} s_{i,j,k} \cdot\tilde{\boldsymbol{\psi}}_{i,j,k}\Bigr)
\end{align}
\vspace{0.5mm}
\end{minipage}}
where the reference coefficients $\{s_{*,*,k}\}$ are predicted by a light-weighted convolution head.  

\noindent\textbf{\ourpe~for Sparse 3D Detectors}
As shown in Fig.~\ref{fig:pe}, \ourpe~can be easily plugged into any PETR-style detectors~\cite{liu2022petr,liu2022petrv2}. The only modification that we need to make is to compute the positional encoding of the query vectors in a similar way that we compute \ourpe. Specifically, given a 3D query anchor $[x_q, y_q, z_q]^{\top}$ in the LIDAR coordinate frame, we compute its positional encoding by:
\resizebox{1.0\linewidth}{!}{
  \begin{minipage}{1.2\linewidth}
\begin{align}
    \boldsymbol{\psi}_{q} &= \mathrm{MLP}\Bigl(\mathrm{Concat}\bigl( \xi(d_{q}), \xi(\sin{\theta_{q}}), \xi(\cos{\theta_{q}}), \xi(z_{q}) \bigr)\Bigr) \\
    &d_{q} = \sqrt{x_{q}^2 +y_{q}^2},~\sin{\theta_{q}} = \frac{y_{q}}{d_{q}},~\cos{\theta_{q}} = \frac{x_{q}}{d_{q}}
\end{align}
\end{minipage}}

\subsection{BEV Transformation with \ourmethod}
\label{sec:bevTransformer}

Empowered by the aforementioned \ourpe, we introduce \ourmethod, an efficient transformer module for BEV view transformation. Our model's overview is shown in Fig.~\ref{fig:pipeline}.

\noindent\textbf{Problem Description.} The input of \ourmethod~is the multi-view image features $\mathbf{F}^I \in \mathbb{R}^{N_c \times H_I \times W_I \times C}$ and the camera parameters. The output is a unified BEV representation $\textbf{F}^B \in \mathbb{R}^{H_B \times W_B \times C}$ where $H_B$ and $W_B$ are the height and width of the BEV features, which are used to solve different downstream tasks, e.g., 3D object detection.

\noindent\textbf{BEV Queries.}
Similar to how we compute positional encodings for the image features, \ourmethod~computes the BEV query vectors using the center coordinates of each BEV grid. Specifically, for the grid with center coordinates $[x, y]^{\top}$, we compute its BEV query vector by:
\resizebox{1.0\linewidth}{!}{
  \begin{minipage}{1.1\linewidth}
\begin{align}
 \boldsymbol{q}^B &= \mathrm{MLP}\Bigl(\mathrm{Concat}\bigl( \xi(d), \xi(\sin{\theta}), \xi(\cos{\theta}) \bigr)\Bigr) \\
    &d^B_{i,j} = \sqrt{x^2 +y^2},~\sin{\theta} = \frac{y}{d},~\cos{\theta} = \frac{x}{d}
\end{align}
\end{minipage}}
Different from sparse 3D detectors, here we remove the height component when encoding the BEV positional information. As a consequence, the features of different heights will be automatically aggregated. 

\noindent\textbf{Width Features as Keys \& Values.}
Intuitively, the BEV features can be easily computed by querying the multi-view image features, added with 3D positional encodings, with the BEV query vectors. However, as the sizes of BEV feature maps are usually in high resolution, e.g., 128$\times$128, such a naive approach will post a significant computational overhead, severely limiting the efficiency and scalability of the model. To overcome this difficulty, we follow recent works~\cite{zhou2022matrixvt} to pool the image features $\mathbf{F}^I$ in the vertical direction, yielding width features $\mathbf{F}^W \in \mathbb{R}^{N_c \times W_I \times C}$ to serve as the attention keys and values. Compared with the full image features $\mathbf{F}^I$, the width features are $H_I$ times smaller, greatly reducing the computational cost. 

\noindent\textbf{Adapting \ourpe~for Width Features.}
In Sec.~\ref{sec:pe}, we described how we compute \ourpe~for the full multi-view features. As shown in Fig.~\ref{fig:pe}, to compute \ourpe~for the width features, we use a light-weighted \textit{Conv head} to predict a discrete categorical height distribution~\cite{zhou2022matrixvt} over each image column, based on which the positional encoding for a width feature is computed by aggregating the pixel-wise positional encodings on its corresponding column. Specifically, for the $j$-th width feature and the predicted height distribution $\{t_{i,j}|~i\in |H_I|\}$, we compute its \ourpe~by:
\begin{align}
    & \boldsymbol{\psi}_{j} = \mathrm{MLP}\Bigl(\sum_{i=1}^{H_I} t_{i,j} \cdot \tilde{\boldsymbol{\psi}}_{i,j}\Bigr),~~\mathrm{where}~~\sum_{i=1}^{H_I} t_{i,j}=1
\end{align}
Similar to the BEV queries, we remove the height component when computing the pixel-wise positional encodings. 

\noindent\textbf{Computing BEV Representations.}
After acquiring the BEV queries $\mathbf{Q}^B$, width features $\mathbf{F}^W$ and \ourpe~for the width features $\boldsymbol{\Psi}^W$, \ourmethod~uses a single transformer decoder layer to compute the BEV features $\mathbf{F}^B$: 
\begin{align}
   \mathbf{U}^B &= \mathbf{Q}^B + \mathrm{MHA}(\mathbf{Q}^P, \mathbf{F}^W+\boldsymbol{\Psi}^W, \mathbf{F}^W)\mathrm{,}\\
   \mathbf{F}^B &= \mathbf{U}^B + \mathrm{FFN}(\mathbf{U}^B)
\end{align}
$\mathrm{MHA}$ and $\mathrm{FFN}$ are the multi-head attention operation and the feed-forward network in a transformer layer~\cite{Transformer_NIPS2017_Vaswani}. Note that we removed the self-attention module in the standard transformer decoder layer~\cite{Transformer_NIPS2017_Vaswani} as it is computationally expensive and has no impact on model performance.

\subsection{Refining Width Features}
\label{sec:refinement}

Compressing 2D image features into 1D width features can greatly improve model efficiency and scalability. However, important information may also be lost during compression. Therefore, we introduce the following two techniques to compensate for any potential information loss.

\noindent\textbf{\refineFormer.}
\refineFormer~is a light-weight transformer decoder~\cite{Transformer_NIPS2017_Vaswani}. It refines an initial width feature by making it attend to and retrieve information from both other width features and the original image features. As illustrated in Fig.~\ref{fig:pipeline}, the initial width feature is computed by MaxPooling the height dimension of the image features. In \refineFormer, a width feature first retrieves information from other width features with a self-attention operation; it then retrieves information from its corresponding image column  with a cross-attention operation. Finally, a feed-forward network is used to compute the final width feature. \refineFormer~is highly efficient because it has a linear complexity with respect to the input image size. Specifically, for an image feature map with size $(H,W)$, the computational complexity is {\small $O(W^2)+O(WH)$}. In Sec.~\ref{sec:exp_ablation}, we show that adding \refineFormer~only incurs minimal latency to the model.

\noindent\textbf{Complementary Tasks.}
To further improve the representation ability of width features, during training, we train the model with complimentary tasks to directly inject task-related information into the width features, which is motivated by BEVFormer v2~\cite{yang2022bevformer2}. As shown in Fig.~\ref{fig:pipeline}(a), the complementary tasks include a monocular 3D detection task and a height prediction task. Specifically, we append an FCOS3D-style head~\cite{wang2021fcos3d} for the two tasks. The head takes 1D width features as input and detects 3D objects in a monocular manner. To make it able to take 1D width features as input, we made two modifications: (1) We changed all 2D convolution operations to 1D convolutions; (2) we ignored the height range and only limited the width ranges during label encoding. To align the complementary tasks with \ourmethod, we changed the original regressional depth estimation to a categorical style. For height prediction, we appended an extra branch to the FCOS3D head to predict an object's height location in the original image features, which can complement the information lost in height pooling. Note that the head for the complementary tasks can be completely removed during model inference, so it will not harm inference efficiency. Also, training the auxiliary head only consumes $<$10M extra GPU memory, so it has minimal effect on training efficiency.

\section{Experiments}
\label{sec:exp}
\subsection{Experiment Settings}

\begin{table}[!t]
\begin{center}
\resizebox{\linewidth}{!}{
\begin{tabular}{l|l|cc|ccccc}
\hline
Detector & PE & mAP & NDS & mATE & mASE & mAOE & mAVE & mAAE \\
\hline
\multirow{5}{*}{PETR-DN} & PETR & 34.3 & 37.2 & 77.0 & 27.7 & 59.3 & 112.6 & 35.2  \\
      & FPE & 35.1 & 37.5 & 76.5 & 27.7 & 60.4 & 126.3 & 35.7  \\
      & CAPE & 34.7 &  40.6 & - & - & - & - & - \\
      & Ours & \textbf{37.1} & \textbf{41.2} & \textbf{72.3} & \textbf{27.5} & \textbf{54.8} & \textbf{94.1} & \textbf{24.3}\\
\hline
\multirow{4}{*}{StreamPETR} & PETR & 37.2 & 47.7 &69.1 & 27.9 & 62.5 & 28.9 & 20.4 \\
      & FPE & 38.6 & 48.7 & 66.1& \textbf{27.4} & 63.0 & 28.3 & 20.5 \\
      & 3DPPE & 26.4 & 41.3 & 78.3 & 28.2 & 61.3 & 31.5 & \textbf{19.8} \\
      & Ours & \textbf{40.2} & \textbf{49.9} & \textbf{64.5} & 27.8 & \textbf{60.8} & \textbf{27.6} & 20.5 \\
\hline
\end{tabular}}
\end{center}
\vspace{-4mm}
\caption{\small Comparison of different positional encoding methods using PETR-DN, and StreamPETR detectors. ResNet-50-DCN is used as the default backbone. The input size is set to 512$\times$1408 for PETR-DN and 256$\times$704 for StreamPETR. All models are trained for 24 epochs without CBGS.}
\label{table:sparseCompare}
\vspace{-2mm}
\end{table}

\begin{table}[!t]
\begin{center}
\resizebox{\linewidth}{!}{
\begin{tabular}{l|l|cc|ccccc}
\hline
Detector & VT Method & mAP & NDS & mATE & mASE & mAOE & mAVE & mAAE \\
\hline
\multirow{6}{*}{BEVDet} & IPM & 25.3 & 34.5 & 78.5 & 27.6 & 62.5 & 85.9 & 26.6 \\
      & LSS & 29.5 & 37.1 & 73.9 & \textbf{27.3} & 61.2 & 88.1 & \textbf{24.8} \\
      & MatrixVT & 28.9 & 36.5 & 74.6 & 28.3 & \textbf{60.0} & 89.5 & 27.3 \\
      & FastBEV & 28.9 & 37.1 & 73.3 & 28.1 & 62.6 & \textbf{82.6} & 27.1 \\
      & BEVFormer & 29.1 & 34.1 & 76.1 & 28.3 & 71.8 & 97.2 & 30.0 \\
      & \ourmethod & \textbf{30.7} & \textbf{37.3} & \textbf{72.8} & 27.6 & 63.7 & 89.6 & 26.8 \\
\hline 
\multirow{6}{*}{BEVDet4D} & IPM & 27.1 & 41.0 & 77.8 & 28.6 & 57.9 & 39.7 & 21.5  \\
      & LSS & 32.8 & 45.7 & 71.0 & 27.9 & 51.2 & 36.0 & 20.5  \\ 
      & MatrixVT & 32.4 &  45.8 & \textbf{69.6} & \textbf{27.6} & \textbf{51.9} & 36.3 & \textbf{18.9} \\
      & FastBEV & 30.8 & 42.4 & 73.7 & 28.1 & 53.7 & 51.6 & 22.4 \\
      & BEVFormer & 31.1 & 41.1 & 74.9 & 28.2 & 63.7 & 52.6 & 24.0 \\
      & \ourmethod & \textbf{34.0} & \textbf{46.3} & 70.4 & 27.9 & 52.9 & \textbf{35.4} & 19.7 \\
\hline
\end{tabular}}
\end{center}
\vspace{-4mm}
\caption{\small Comparison of different BEV view transformation methods using BEVDet and BEVDet4D detectors. ResNet-50 is used as the default backbone network. The input size is set to 256$\times$704. All models are trained for epochs with CBGS.}
\label{table:baselineCompare}
\vspace{-6mm}
\end{table}

\noindent\textbf{Dataset.}
We benchmark our method on the commonly used nuScenes dataset~\cite{caesar2020nuscenes}, which includes 700, 150 and 150 scenes for training, validation and testing. Each scene contains 6-view images that cover the whole surrounding environment. We follow the official evaluation protocol. Specifically, for the 3D object detection task, except for the commonly used  mean average precision (mAP), the evaluation metrics also include the nuScenes true positive (TP) errors, which contain mean average translation error (mATE), mean Average Scale Error (mASE), mean Average Orientation Error (mAOE), mean average velocity error (mAVE) and mean average attribute error (mAAE). We also report the nuScenes detection score (NDS). We use the nuScenes training set for model training and evaluate the models on the nuScenes \textit{val} set.

\noindent\textbf{Implementation Details.}
We test our \ourpe~for sparse 3D object detectors following the open-sourced implementation in~\cite{wang2023exploring}. We test our proposed \ourmethod~using two 3D detection architectures: BEVDet~\cite{huang2021bevdet} and BEVDet4D~\cite{huang2022bevdet4d}.  which covers single-frame and multi-frame settings. We adopt the implementation of all three detectors in the BEVDet~\cite{huang2021bevdet} code base. Unless otherwise specified, we use BEVDet's~\cite{huang2021bevdet} default data pre-processing and augmentation settings. We also follow~\cite{huang2021bevdet} to set the BEV feature size to $128\times 128$ and BEV channel size to $64$. For BEVDet4D~\cite{huang2022bevdet4d} and BEVDepth4D~\cite{li2022bevdepth} experiments, we follow the original BEVDet4D implementation where only one history frame is used. All models are trained for 24 epochs with CBGS~\cite{zhu2019cbgs}. An ImageNet pre-trained ResNet-50 is used as the default backbone network. All training and CUDA latency measurements are conducted using NVIDIA 3090 GPUs.

\subsection{Main Results}
\label{sec:exp_mainResults}

\begin{figure}[!t]
  \begin{center}
    \includegraphics[width=0.8\linewidth]{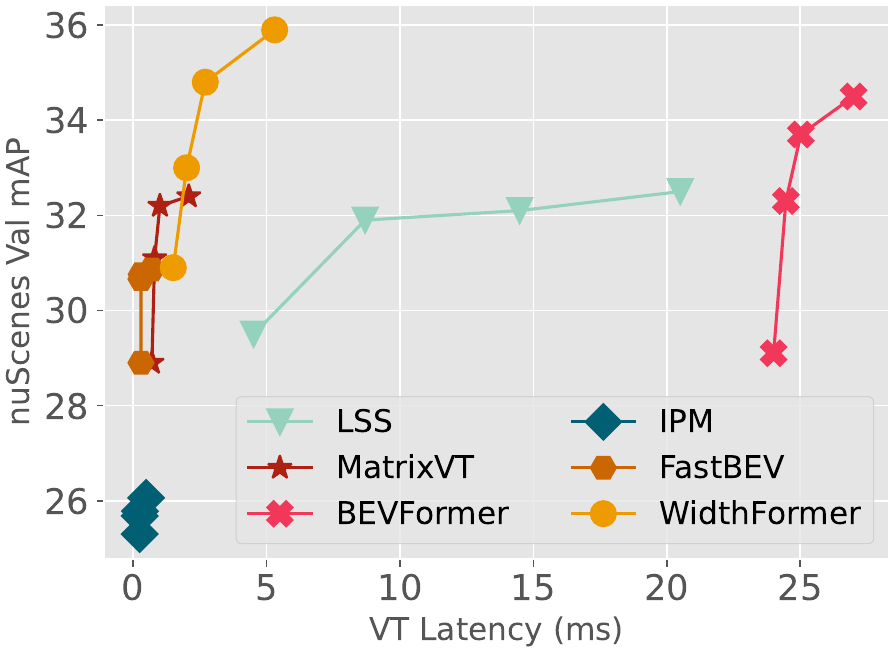} 
  \end{center}
  \vspace{-0.15in}
    \caption{\small CUDA Latency and mAP trade-off comparison of different VT methods using BEVDet on various size settings.}
    \vspace{-7mm}
    \label{fig:baselineScale}
\end{figure}

\noindent\textbf{\ourpe~vs Other Positional Encoding.}
In Tab.~\ref{table:sparseCompare}, we compare our \ourpe~with other 3D positional encoding methods using a baseline detector PETR~\cite{liu2022petr} and the state-of-the-art StreamPETR~\cite{wang2023exploring}. The competing PE approaches include the original 3D PE proposed in the PETR paper~\cite{liu2022petrv2}, the featurized 3D PE~\cite{liu2022petrv2}, CAPE~\cite{Xiong2023CAPE} and the recently proposed 3DPPE~\cite{shu20223d}. For a fair comparison, all models are trained for 24 epochs without CBGS using ResNet-50-DCN as the backbone networks.  The results show that our \ourpe~achieve the best performance on all three baselines. Specifically, on the PETR-DN detector, our method achieves significant 2.8 and 4.0 improvements on mAP and NDS over the baseline PETR, respectively. Also, on the high-performing StreamPETR, \ourpe~improves the mAP by 1.6 and NDS by 1.2 over the baseline StreamPETR that uses FPE as positional encoding. We also notice that, under our training recipe, the recently proposed 3DPPE does not achieve a satisfactory performance, showing its sensitivity to training setups. These results validate our \ourpe's effectiveness.

\noindent\textbf{\ourmethod~vs Other VT Methods.}
In Tab.~\ref{table:baselineCompare}, we compare our proposed \ourmethod~with other VT methods using the BEVDet~\cite{huang2021bevdet} and BEVDet4D~\cite{huang2022bevdet4d} 3D detectors. The competing methods include (1) Inverse Perspective Mapping (IPM)~\cite{8814050}, (2) Lift-splat based LSS~\cite{philion2020lift}, MatrixVT~\cite{zhou2022matrixvt} and FastBEV~\cite{li2023fastbev}, and (3) transformer-based BEVFormer~\cite{bevformer} (6-layers version). Note that the competing methods may use different backbone \& head settings, multi-frame fusing strategies and training recipes in their original paper. Here, to ensure a fair comparison, we only use those VT methods to compute BEV representations and kept all other settings same following the original 3D detector~\cite{huang2021bevdet,huang2022bevdet4d}. We also made sure the BEV representations computed from different VT methods have the same resolution and channel dimensions. Our method achieves better mAP and NDS performance than all competing methods on all three baseline architectures. Note that our method and BEVFormer~\cite{bevformer} are both transformer-based, but our achieves better performance than BEVFormer in both single-frame and multi-frame settings. In addition, MatrixVT~\cite{zhou2022matrixvt} and \ourmethod~both compute BEV representations from width features, but ours is consistently better than MatrixVT on both settings. These results validate the effectiveness of our method as a general VT approach. In Fig.~\ref{fig:baselineScale}, we compare the CUDA latency and detection mAP curve of our method and other VT methods with different input resolutions and feature channels ( $\{H_I,W_I,C\}$) on the BEVDet detector. Specifically, we tested four settings {$\{256,704,64\}$, $\{384,1056,64\}$, $\{512,1408,64\}$ and $\{512,1408,128\}$}. The result demonstrates the good speed and accuracy balance of our model: the detection mAP keeps improving as the inputs scale-up, while the VT latency keeps low. We do notice that IPM, MatrixVT and FastBEV achieve a better speed than ours, but this is at the expense of accuracy.

\begin{table}[t]
  \begin{center}
    \resizebox{\linewidth}{!}{
    \begin{tabular}{l@{\hskip 40pt}|c@{\hskip 8pt}c|c}
    \hline
    Setting & mAP &NDS & VT Latency \\ 
    \hline
    LSS Baseline & 29.5 & 37.1 & 4.5 ms \\ 
    + Transformer & - & - & - \\
    + \ourpe & \textbf{31.0} & \textbf{38.0} & 4.6 ms \\
    + Width Feature & 28.3 & 35.3 & 1.3 ms \\
    + Refine Transformer & 29.3 & 36.2 & 1.5 ms \\
    + Auxiliary Head & 30.7 & 37.3 & \textbf{1.5 ms} \\
    \hline 
    \end{tabular}}
  \end{center}
  \vspace{-0.4cm}
  \caption{\small {Step-by-step ablation studies.}}
  \vspace{-2mm}
\label{tbl:attnAblation}
\end{table}

\subsection{Ablation Studies and Discussions}
\label{sec:exp_ablation}

We conduct ablation studies to test different design choices of our method. Unless otherwise specified, the experiments are conducted using a ResNet-50-based BEVDet.

\noindent\textbf{Building \ourmethod~step by step.}
We first examine the effect of our proposed modules by gradually adding them to the model and reporting accuracy \& speed results in Tab.~\ref{tbl:attnAblation}. We start with the original LSS~\cite{philion2020lift} baseline that achieves 29.5 mAP with 4.5 ms VT latency. We then replace LSS with a transformer layer, where BEV queries directly interact with the image features without 3D positional encoding (PE). However, the model fails to converge. We then add our \ourpe~encoding to the model, in which we compute PEs for every feature pixel without averaging them in the height dimension. The model achieves 31.0 mAP, which is better than the LSS baseline. This result validates the necessity of 3D PEs. However, querying the whole image features is inefficient (4.6 ms). By adopting width features as attention keys and values, the latency reduces to 1.3 ms. However, the accuracy falls to 28.3 mAP, implying that the feature compression is causing us to lose information. We mitigate this by adding the Refine Transformer to the model, which brings 1.0 mAP improvement with minor computation overhead (+0.2 ms). Finally, we add the complementary tasks that improve the final mAP to 30.7. These results show that the proposed Refine Transformer and complementary tasks can indeed mitigate the information loss problem.

\begin{table}[t!]
\renewcommand\arraystretch{1.0}
  \footnotesize 
  \centering
  \begin{threeparttable}
  \resizebox{\linewidth}{!}{
    \begin{tabularx}{1.0\linewidth}{l@{\hskip 5pt}|CC|C}
    \hline
    Refinement & mAP & NDS & VT Latency \\ 
    \hline
    None & 30.0 & 36.5 & \textbf{1.3 ms} \\
    \hline
    Conv & 30.2 & 36.3 & 1.4 ms \\
    RF & 30.7 & 37.3 & 1.5 ms \\
    Conv+RF & \textbf{30.9} & \textbf{37.2} & 1.6 ms \\
    \hline 
    \end{tabularx}}
    \end{threeparttable}
  \vspace{-0.1cm}
  \caption{\small Ablation study on feature refinement strategies.}
\label{tbl:ablationRefine}
\vspace{-6mm}
\end{table}

\begin{figure}[!t]
  \begin{center}
    \includegraphics[width=0.8\linewidth]{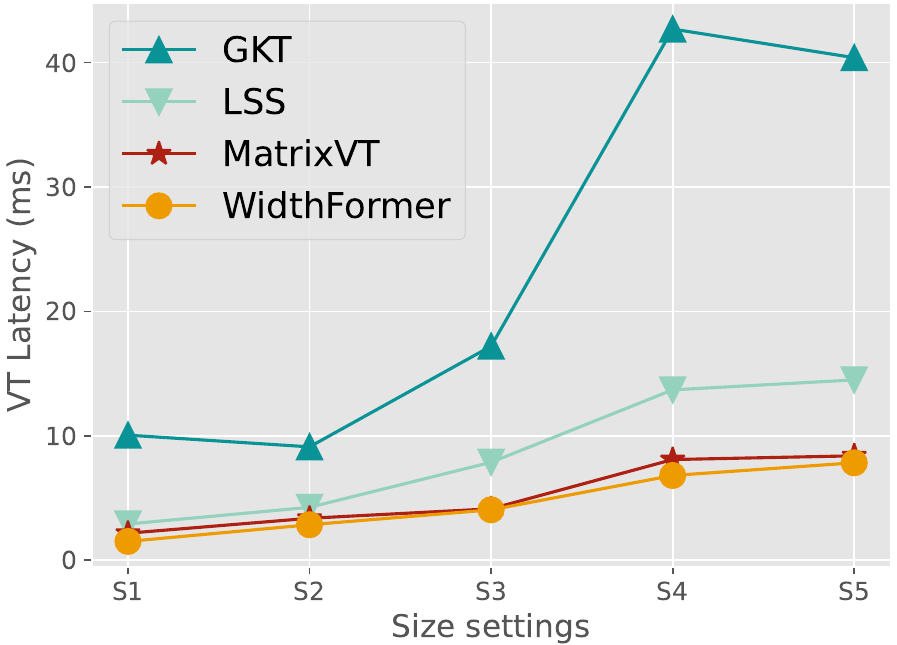} 
  \end{center}
  \vspace{-4mm}
    \caption{\small Latency comparison of different VT methods on the Horizon Journey-5 computation solutions.}
    \vspace{-3mm}
    \label{fig:j5}
\end{figure}

\begin{table}[t!]
\renewcommand\arraystretch{1.0}
  \footnotesize 
  \centering
  \begin{threeparttable}
  \resizebox{\linewidth}{!}{
    \begin{tabularx}{1.0\linewidth}{l@{\hskip 12pt}|C@{\hskip 8pt}C}
    \hline
    Aux Loss & mAP &NDS  \\ 
    \hline
    None & 29.3 & 36.2  \\ 
    \hline
    + Width FCOS3D & 29.7 & 36.4  \\
    + Height Loss & 30.5 & 36.9 \\
    + CateDepth & \textbf{30.7} & \textbf{37.3} \\
    \hline 
    \end{tabularx}}
    \end{threeparttable}
  \vspace{-0.1cm}
    \caption{\small Ablation Study on the auxiliary head and its losses.}
\label{tbl:ablationFcos}
\vspace{-5mm}
\end{table}

\begin{figure*}[!t]
    \centering
    \includegraphics[width=0.8\textwidth]{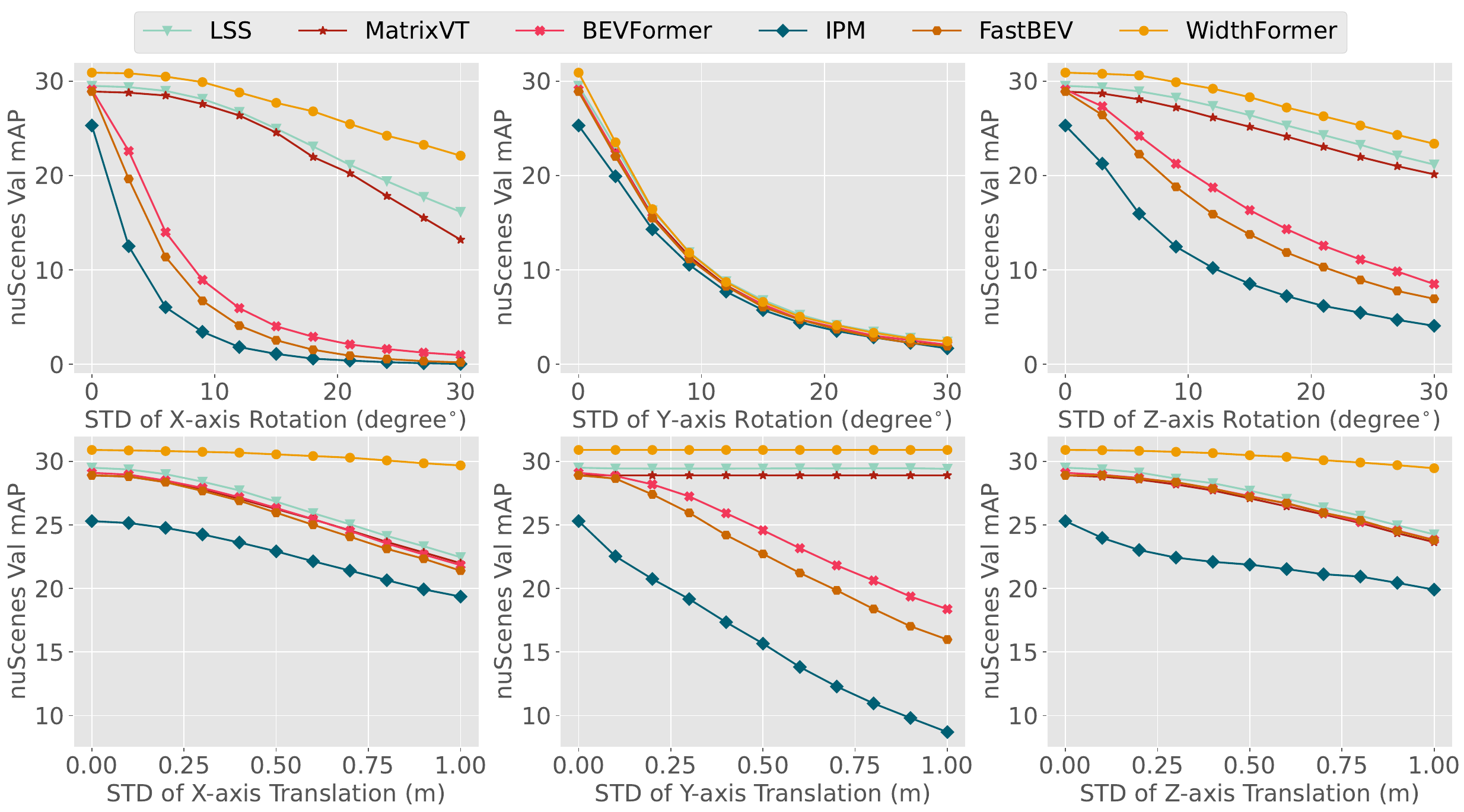}
    \vspace{-0.1cm}
    \captionsetup{font={small}}
\caption{Comparison of VT method's robustness against different types of camera perturbations. We perturb each camera independently.}
\label{fig:baselineRobust}
\vspace{-1mm}
\end{figure*}

\begin{table*}[t!]
  \begin{center}
  \resizebox{\textwidth}{!}{
  \begin{tabular}{l@{\hskip 5pt}|c|c|c|c|cc|ccccc}
\hline
Method & Backbone & Resolution & Modality & MF & mAP~$\uparrow$ & NDS~$\uparrow$ & mATE~$\downarrow$ & mASE~$\downarrow$ & mAOE~$\downarrow$ & mAVE~$\downarrow$ & mAAE~$\downarrow$ \\

\hline
BEVDet~\cite{huang2021bevdet} & Swin-B & 512$\times$1408 & C &   & 34.9 & 41.7 & 63.7 & 26.9 & 49.0 & 91.4 & 26.8 \\
PETR~\cite{liu2022petr} & Res-101 & 512$\times$1408 & C &   & 35.7 & 42.1 & 71.0 & 27.0 & 49.0 & 88.5 & \textbf{22.4} \\
BEVFormer-S~\cite{bevformer} & Res-101 & 900$\times$1600 & C &   & 37.5 & 44.8 &- & - &- & - & - \\
{BEVDet+\ourmethod} & Res-101 & 512$\times$1408 & C &   & \textbf{37.9} & \textbf{44.8} & \textbf{62.7} & \textbf{26.1} & \textbf{45.1} & \textbf{84.0} & 23.7 \\

\hline 
BEVDet4D~\cite{huang2022bevdet4d} &Swin-B & 640$\times$1600 & C & \checkmark & 39.6 & 51.5 & 61.9 & \textbf{26.0} & 36.1 & 39.9 & 18.9 \\
BEVFormer~\cite{bevformer} & Res-101 & 900$\times$1600 & C & \checkmark & 41.6 & 51.7 & 67.3 & 27.4 & 37.2 & 39.4 & 19.8 \\
PolarFormer-T~\cite{jiang2022polar} & Res-101 & 900$\times$1600 & C & \checkmark & 38.3 & 48.8 & 70.7 & 26.9 & 34.4 & 51.8 & 19.6 \\
Fast-BEV~\cite{li2023fastbev} & Res-101 & 900$\times$1600 & C & \checkmark & 40.2 & 53.1 & 58.2 & 27.8 & \textbf{30.4} & 32.8 & 20.9 \\
PETRv2~\cite{liu2022petrv2} & Res-101 & 640$\times$1600 & C & \checkmark & 42.1 & 52.4 & 68.1 & 26.7 & 35.7 & 37.7 & \textbf{18.6} \\
BEVDepth~\cite{li2022bevdepth} & Res-101 & 640$\times$1600 & C+L & \checkmark & 41.2 & \textbf{53.5} & \textbf{56.5} & 26.6 & 35.8 & 33.1 & 19.0 \\ 
BEVDistill~\cite{chen2023bevdistill}&  Res-101 & 640$\times$1600 & C+L & \checkmark & 41.7 & 52.4 & - & - & - & - & - \\
{BEVDet4D+\ourmethod} & Res-101 & 512$\times$1408 & C & \checkmark & \textbf{42.3} & 53.1 & 60.9 & 26.9 & 41.2 & \textbf{30.2} & 21.0 \\
\hline
CAPE-T~\cite{Xiong2023CAPE} & V2-99 & 320$\times$800 & C & \checkmark & 44.0 & 53.6 & 67.5 & 26.7 & 39.6 & 32.3 & 18.5 \\
StreamPETR~\cite{wang2023exploring} & V2-99 & 320$\times$800 & C & \checkmark & 48.2 & 57.1 & - & - & - & - & - \\
StreamPETR+3DPPE~\cite{shu20223d} & V2-99 & 320$\times$800 & C+L & \checkmark & 49.9 & 58.4 & - & - & - & - & - \\
{StreamPETR+\ourpe} & V2-99 & 320$\times$800 & C & \checkmark & \textbf{49.9} & \textbf{58.5} & \textbf{58.1} & \textbf{25.8} & \textbf{35.8} & \textbf{25.7} & \textbf{19.0} \\

\hline
\end{tabular}}
\end{center}
  \vspace{-0.4cm}
   \caption{\small Scaling-up detection results and comparison with other state-of-the-art 3D object detectors on nuScenes \textit{val} set. `MF' stands for multi-frame fusion; `C' stands for camera; `L' stands for LIDAR.}
  \label{tab:sota}
  \vspace{-6mm}
\end{table*}

\noindent\textbf{Refine Transformer.} 
We examine the design of our Refine Transformer and present the results in Tab.~\ref{tbl:ablationRefine}. The baseline model that has no refinement to the width features achieves 30.0 mAP and 1.3 ms VT latency. When adding the convolutional-based refinement in MatrixVT~\cite{zhou2022matrixvt}, the accuracy is improved to 30.2 mAP. However, when adding our proposed Refine Transformer, the mAP is greatly improved to 30.7. We notice that our Refine Transformer is slightly slower than the convolution-based refinement but that this is negligible compared to the overall latency. We also tried to combine Refine Transformer with convolution-based refinement but found it only brought a marginal improvement in accuracy. This experiment validates the effectiveness of our Refine Transformer.

\noindent\textbf{Complementary Tasks.} 
We study our proposed complementary tasks and report the results in Tab.~\ref{tbl:ablationFcos}. When training with complementary tasks is disabled, our model achieves 29.3 mAP. Then we add the 3D monocular detection task as in FCOS3D~\cite{wang2021fcos3d}, the detection accuracy is improved to 29.7 mAP. Then we add the height estimation branch---where the model estimates an object's height in the 2D image feature---the performance is further improved by 30.5 mAP. Finally, we modified the regression depth estimation loss into a categorical depth estimation loss, which gave us a final mAP to 30.7. This experiment validates the effectiveness of our training strategy.

\noindent\textbf{Speed on Edge Computing Device.}
In Fig.~\ref{fig:j5}, we report the speed-testing results comparing \ourmethod~and other VT methods on the Horizon Journey-5~\cite{j5_website}. We compare our method with three other competing methods: GKT~\cite{GeokernelTransformer}, LSS~\cite{philion2020lift} and MatrixVT~\cite{zhou2022matrixvt}. We report the VT latency with different size settings {$\{H_I,W_I,C,H_B\}$} using square BEV grids. Specifically, the five size settings are: {S$_1$=$(128,352,64,128)$, S$_2$=$(256,704,64,128)$, S$_3$=$(256,704,128,128)$, S$_2$=$(256,704,128,192)$ and S$_5$=$(304,832,128,192)$}. Note that the Journey-5 has special support for LSS, making it much faster than naive implementations. The results show that \ourmethod~is nearly two times faster than LSS and several times faster than GKT. We also notice that MatrixVT has a similar latency to ours, which differs from the results in Fig.~\ref{fig:baselineScale}. After investigation, we found that MatrixVT's onboard speed is severely limited by the two very large ray and ring matrices. On the other hand, as shown in other experiments, \ourmethod~has a better accuracy (Tab.~\ref{table:baselineCompare}) than MatrixVT. Therefore, we argue that ours is a better choice for the actual autonomous driving applications.

\noindent\textbf{Robustness under camera perturbation.} In this section, we study the robustness of our method and other VT approaches under 6DoF camera perturbation: unexpected rotation and translation w.r.t. $X$, $Y$ and $Z$ axis in the camera coordinate system. Specifically, the perturbations are implemented by modifying the camera extrinsics with zero-mean Gaussian noises, whose magnitude is controlled by the standard deviation of the added Gaussian. Specifically, for the $n$-th camera, the noised rotation matrix is computed by $\mathbf{\hat{R}}^{n} = \mathbf{R}^{n} * \mathbf{R}'$, and the noised translation vector is computed by $\mathbf{\hat{T}}^{n} = \mathbf{T}^{n} + \mathbf{T}'$, where $*$ denotes matrix multiplication; $\mathbf{R}'$ and $\mathbf{T}'$ are the noisy rotation matrix and translation vector. We then use $\mathbf{\hat{R}}^{n}$ or $\mathbf{\hat{T}}^{n}$ to compute Eq.~\ref{eq:extrinsic}. We show the different VT methods' mAP curves w.r.t. different types of perturbation in Fig.~\ref{fig:baselineRobust}. Interestingly, we find that the different VT methods are sensitive to different types of perturbations. For example, while the $Y$-axis translation causes performance decrease to IPM and BEVFormer, our method and Lift-splat based LSS and MatrixVT are completely insensitive to this perturbation because the height dimension is omitted in these approaches. In general, we observe that the Lift-splat based approaches are more robust than IPM and BEVFormer. The reason is that IPM and BEVFormer rely on projecting 3D space coordinates to 2D images, and a small amount of perturbation may cause a huge difference in the projected location, therefore harming performance. Particularly, our method is very robust to the translation perturbations with minor performance degrades compared to all competing approaches. Our method is also robust against $X$ and $Z$ axis rotation. However, we notice that all VT methods, including ours, have similar strong sensitivities to the $Y$-axis rotation, with the mAPs rapidly decreasing to zero as the noise magnitudes rise. The reason is that the $Y$-axis rotation can severely damages the predefined multi-view arrangement~\cite{caesar2020nuscenes}.

\noindent\textbf{Scaling-up Experiments.}
In Tab.~\ref{tab:sota}, we scale up our model and compare it with other state-of-the-art 3D detection approaches on the nuScenes $\textit{val}$ set. Specifically, for BEV-based detectors, we use ResNet-101 as the image encoder; we also scale up the input resolution and BEV channel dimensions to 512$\times$1408 and 128, respectively. The results show that our method achieves on-par performance with other state-of-the-art methods. For example, BEVFormer~\cite{bevformer} achieves 37.5 mAP \& 44.8 NDS and 41.6 mAP \& 51.7 NDS in single-frame and multi-frame settings, while ours achieves 37.9 mAP \& 44.8 NDS and 40.1 mAP \& 52.7 NDS. Additionally, the superior performance of StreamPETR+\ourpe~ further validates the effectiveness of our design.

\section{Conclusion}

We introduce \ourmethod, a new transformer-based BEV transformation method. With the help of a new 3D positional encoding \ourpe, which can also be used to boost a sparse 3D detector's performance, \ourmethod~use a single layer of transformer decoder to compute the BEV representations from the vertically compressed visual features. Compared with previous BEV transformation approaches, our model is more efficient and easy to deploy to edge-computing devices. We also show that \ourmethod~holds a good robustness against camera perturbations.

{\small
\bibliographystyle{ieee}
\bibliography{bib}
}

\clearpage

\end{document}